\documentclass{article}
\usepackage{amsmath}
\usepackage[colorinlistoftodos, textsize=small]{todonotes}
\usepackage{float}
\usepackage{graphicx}
\usepackage{subcaption}
\usepackage{multirow}
\usepackage{booktabs}
\usepackage{graphicx}
\usepackage[table]{xcolor}
\usepackage{nicematrix}
\usepackage{todonotes}
\usepackage{wrapfig}
\usepackage[font=footnotesize]{caption}
\usepackage{subcaption}
\usepackage{amssymb}

\definecolor{daimonblue}{HTML}{2B6CB0}
\definecolor{efleshred}{HTML}{C53030}
\definecolor{flexitacgreen}{HTML}{2F855A}

\newcommand{\Daimon}{\textcolor{daimonblue}{Daimon}}
\newcommand{\eFlesh}{\textcolor{efleshred}{eFlesh}}
\newcommand{\FlexiTac}{\textcolor{flexitacgreen}{FlexiTac}}
\usepackage[preprint]{corl_2026} 

\title{\textsc{TactX}: Learning Shared Tactile Representations Across Diverse Sensors}

\author{
  Junsung Park$^{1,2*}$, 
  Sachin Bhadang$^{1*}$,
  Carmelo Sferrazza$^{3}$,
  Sha Yi$^{1}$,
  Xiaolong Wang$^{1}$ \\[0.8em]
  $^{1}$UC San Diego,
  $^{2}$Seoul National University,
  $^{3}$Amazon FAR \\[0.8em]
  $^{*}$Equal contribution; author order determined by coin flip.
}

\begin{document}

\maketitle

\vspace{-2.7em}
\begin{center}
\small
Homepage: \url{https://tactx-project.github.io/}
\end{center}

\vspace{1.2em}

\vspace{-20pt}
\begin{figure}[h]
    \centering
    \includegraphics[width=1\textwidth]{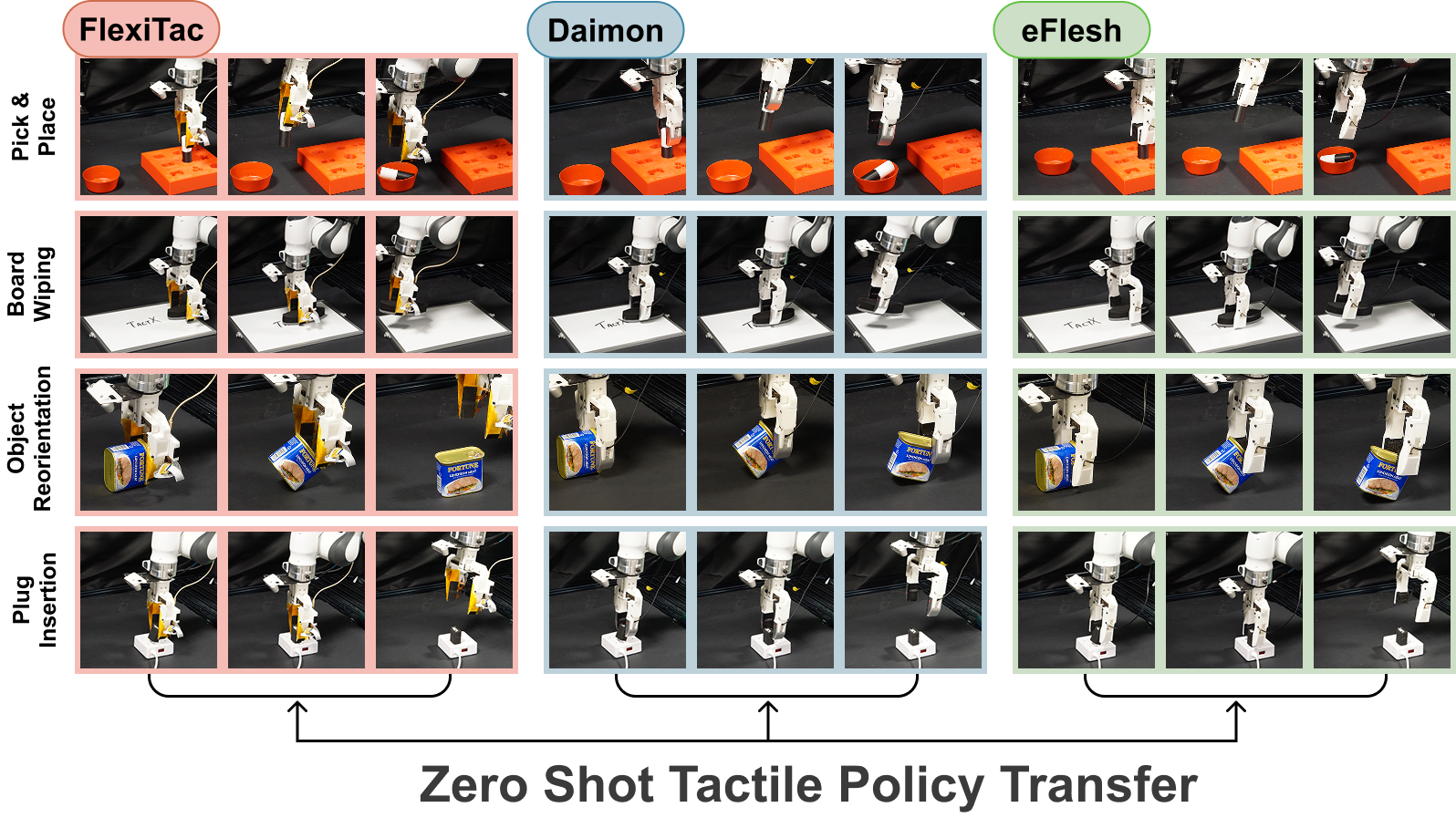}
    \caption{
    \textsc{TactX} learns a shared latent representation that aligns heterogeneous tactile sensors and enables zero-shot transfer of tactile-conditioned policies.
    }
    \label{fig:teaser}
\end{figure}


\begin{abstract}
Tactile sensors provide critical information for contact-rich manipulation, yet tactile representations and policies remain tightly coupled to each specific sensor, limiting transferability across robots and hardware platforms. We propose \textsc{TactX}, a framework for learning a transferable tactile representation across sensors spanning three fundamentally different transduction modalities: resistive, magnetic, and vision-based. \textsc{TactX} maps heterogeneous tactile observations into a shared latent space through modality-specific encoders trained on paired contact data. Such paired interactions provide a natural alignment signal across modalities, and the encoders are jointly trained across all sensor pairs, inducing a consistent latent space for all sensor types. Our experiments show that \textsc{TactX} aligns tactile representations across sensors while preserving object-level contact information, as evidenced by sensor-identity prediction and object classification in the learned latent space. We evaluate \textsc{TactX} on four contact-rich manipulation tasks---pick-and-place, plug insertion, board wiping, and object reorientation---and show that policies trained with one sensor transfer zero-shot to physically distinct sensors through the shared latent. This improves the average success rate from 27.5\% for vision-only policy to 45.9\%, providing a step toward sensor-agnostic tactile manipulation.
\end{abstract}

\newpage
\section{Introduction}
\label{sec:introduction}

\vspace{-5pt}
Contact-rich manipulation depends on information that vision alone cannot provide. Tactile sensing captures forces, slips, and contact geometries at the robot--world interface, and prior work has shown that it is essential for robust manipulation in cluttered, occluded, or precision-critical settings~\cite{calandra2018,inhand_placeholder,insertion_placeholder,nonprehensile}. However, sensors differ not only in form factor, but also in how they physically measure contact, including through optical deformation, magnetic-field changes, and resistive pressure responses. This diversity has enabled specialized capabilities, but it also makes tactile representations and policies highly hardware-dependent. A policy trained with one tactile sensor is usually tied to that sensor's observation space, and replacing the sensor often requires collecting new demonstrations and retraining the downstream policy.

Cross-sensor tactile learning aims to reduce this dependence, but existing work has largely focused on transferring between vision-based tactile sensors~\cite{unitouch,t3,anytouch,sparsh,rodriguez2025cross}. Calibration or direct transfer across sensors is often insufficient because the same contact can produce substantially different signal distributions on different hardware. The more general question is how tactile representations can be shared across sensors that measure contact through different physical modalities, such as vision-based tactile images~\cite{gelsight,digit,tactip,ninedtact}, magnetic fields~\cite{uskin,reskin,eflesh}, and resistive pressure maps~\cite{flexitac,papillarray}.

To address this challenge, we propose \textsc{TactX}, a framework for learning sensor-agnostic tactile representations across heterogeneous tactile sensors. \textsc{TactX} learns from paired contact data collected using a gripper with a different tactile sensor mounted on each finger. Each grasp produces paired observations, where the two sensors measure the same contact point. Since multiple sensors cannot be mounted simultaneously, we collect data for each sensor pair and train the encoders jointly across pairs, inducing a globally consistent latent space from pairwise supervision. This pairwise formulation also provides a natural extension for incorporating additional sensors, since new modalities can be connected to the shared space through paired contact data.

\textsc{TactX} combines contrastive alignment with self- and cross-reconstruction: contrastive learning pulls paired contacts from different sensors together in latent space~\cite{unitouch,sparsh,infonce}, while reconstruction encourages the latent to preserve object- and contact-level structure. We evaluate this shared representation through both representation-level analyses and zero-shot policy-transfer experiments. Our results show that \textsc{TactX} aligns three heterogeneous tactile sensors into a common latent space while supporting tactile-conditioned robot policies~\cite{act} trained with one sensor and deployed zero-shot with another across four contact-rich manipulation tasks: pick-and-place, plug insertion, board wiping, and object reorientation.

Our contributions are as follows. First, we present \textsc{TactX}, a framework for learning shared tactile representations across fundamentally different sensing modalities---vision-based, magnetic, and resistive---going beyond prior cross-sensor settings that focus on vision-based tactile sensors. Second, we introduce a pairwise training strategy that uses paired contact data to align all sensor pairs into a globally consistent latent space. Third, we demonstrate the practical utility of this shared latent space for robotic manipulation, showing that tactile-conditioned policies trained with one sensor can be deployed on physically distinct sensors without retraining, improving over vision-only transfer baselines by approximately 20\%.

\section{Related Work}
\label{sec:related}

\paragraph{Contact-Rich Manipulation.}
Tactile feedback has been shown to improve robotic manipulation across a wide range of contact-rich tasks: peg-in-hole and electronics insertion~\citep{insertion_placeholder}, in-hand reorientation ~\cite{qi2023generalinhandobjectrotation} and dexterous manipulation~\citep{inhand_placeholder,lin2024learningvisuotactileskillsmultifingered, dexteritygen, dexndm}, sliding, wiping, and pivoting~\citep{nonprehensile,wiping_placeholder}, cable routing~\citep{cable_placeholder}, and grasping under clutter or occlusion~\citep{calandra2018,grasping_placeholder}. These results collectively establish that tactile sensing provides information that vision alone cannot, particularly for sub-millimeter precision and contact-state estimation under occlusion.

\paragraph{Tactile Sensors and Representations.}
Modern tactile sensors span a wide range of transduction principles, including vision-based~\cite{gelsight,digit,ninedtact,gelslim,daimon,softbubble,densetact}, magnetic~\citep{uskin, reskin, eflesh, anyskin}, resistive and capacitive~\citep{flexitac, papillarray}, and piezoelectric or acoustic~\citep{piezo_placeholder, acoustic_placeholder, vibecheck} designs. Recent work has produced strong per-sensor representations, ranging from sensor-specific encoders trained end-to-end~\citep{calandra2018,tactile_policy_placeholder} to large-scale self-supervised pretraining~\citep{sparshskin, sparshx, unit}. Comparative benchmarks~\citep{sparsh,taco} have shown that the usefulness of tactile information depends strongly on sensor modality, material properties, and the task, motivating representations that generalize across sensors rather than being tied to any single device.

\paragraph{Cross-Embodiment and Cross-Sensor Representations.}
A growing line of work seeks unified representations across heterogeneous hardware. In cross-embodiment robot learning, shared \emph{latent action spaces} enable a single policy to drive multiple dexterous hands or arms~\citep{xlvla, lad, cetransfer, cyclevae, univla}, often via per-embodiment encoder--decoder pairs that map into a common space. In cross-sensor tactile learning, prior work has primarily focused on aligning sensors within a shared sensing substrate---typically the vision-based family, where signals share a common image-like representation~\citep{unitouch, t3, anytouch, sparsh, sensorinvariant, anytouch2,cttp}. A smaller body of work addresses sensors with no shared substrate, either by mapping signals to a unified input format~\citep{uniforce, genforce, unitacnv, transforce} or by aligning distributions without paired data~\citep{tactalign}. \textsc{TactX} extends this direction to three transduction modalities simultaneously---resistive, magnetic, and vision-based---and aligns them in latent space without any per-sensor input transformation.

\section{Methodology}
\label{sec:tactx}

\begin{figure}[tb]
    \centering
    \includegraphics[width=\linewidth]{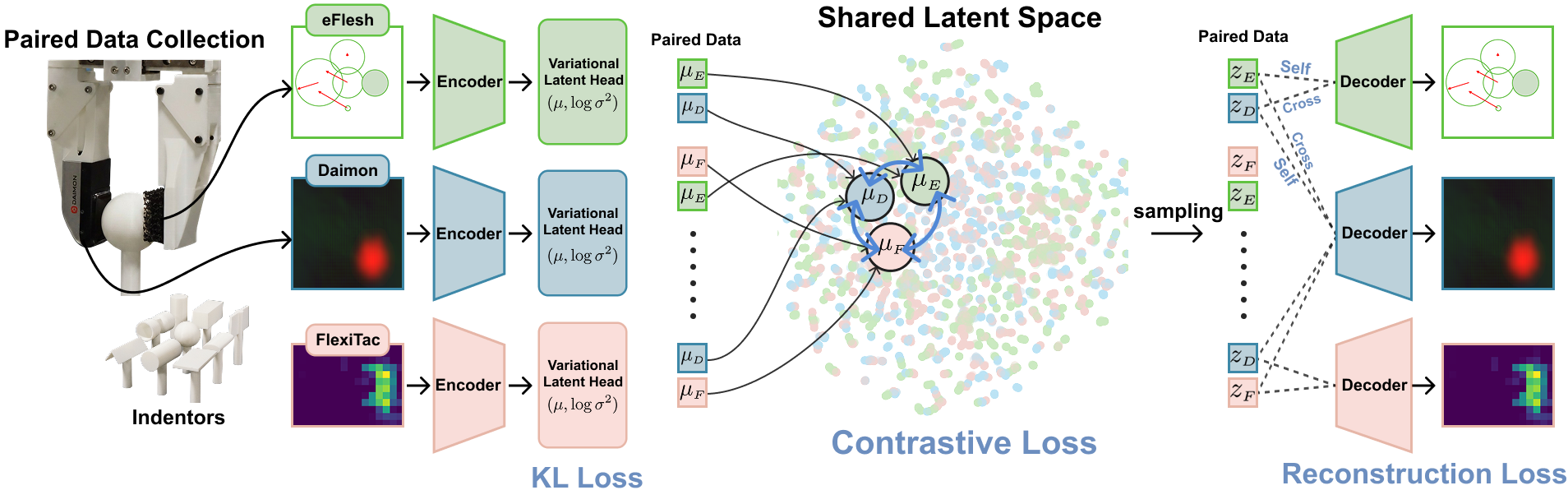}
    \caption{\textbf{\textsc{TactX}} trains on paired contacts from two sensors at a time. Paired observations are encoded into a shared latent space, aligned with InfoNCE, and decoded through self- and cross-reconstruction. Other pairs are trained analogously, yielding a single latent space shared by all three sensors.}
    \label{fig:tactx_architecture}
\end{figure}

Our goal is to align tactile observations from fundamentally different sensing modalities into a shared representation. This is challenging because \textbf{vision-based}, \textbf{magnetic}, and \textbf{resistive} sensors produce observations with different structures, dimensionalities, and sampling rates. \textsc{TactX} addresses this by using modality-specific encoders to map each observation into a shared latent space $\mathcal{Z}$, contrastive learning to align paired contacts across sensors, and reconstruction to preserve contact-relevant information. An overview of the data collection, encoder--decoder architecture, and training losses is shown in Figure~\ref{fig:tactx_architecture}.

\paragraph{Data collection.}
\label{sec:data}

To construct positive pairs across sensors, \textsc{TactX} collects paired contact observations by mounting different tactile sensors on the same gripper. For each sensor pair $S_i, S_j \in \mathcal{S}$, we record quasi-static grasps of rigid symmetric objects, where each grasp provides one measurement from each sensor for the same physical contact. The resulting temporally aligned pairs $(x_i^{(t)}, x_j^{(t)})$ over diverse contact points and 10 indentors form the pair dataset $\mathcal{D}_{ij} = {(x_i^{(t)}, x_j^{(t)})}$. We collect such paired datasets across all sensor pairs, which jointly supervise the sensor-specific encoders. Details on sensor instantiations, hardware, objects, and protocol are provided in Appendix~\ref{app:data}.

\paragraph{Architecture.}
For each sensor $i \in \mathcal{S}$, an encoder $f_i$ maps its native signal $x_i \in \mathcal{X}_i$ through a signal-specific backbone and projection head, which outputs the parameters of a posterior $q_i(z \mid x_i) = \mathcal{N}(\mu_i(x_i), \mathrm{diag}(\sigma_i^2(x_i)))$ over the shared latent $z \in \mathcal{Z} \subset \mathbb{R}^{16}$. The low-dimensional latent encourages the encoders to learn shared contact features while compressing sensor-specific detail. Each sensor has its own decoder $g_i : \mathcal{Z} \to \mathcal{X}_i$, used for both self- and cross-reconstruction. All encoders are trained from scratch such that modalities start from equal footing. Per-sensor backbones and decoder architectures are given in Appendix~\ref{app:arch}.

\paragraph{Forward pass.}
A single training example is one left--right pair $(x_i, x_j)$ from $\mathcal{D}_{ij}$: the signal $x_i$ from the sensor on one finger and $x_j$ from the sensor on the other, both observing the same contact. The two encoders produce posteriors $q_i(z_i \mid x_i)$ and $q_j(z_j \mid x_j)$; we align the posterior means $\mu_i, \mu_j$ in the shared latent (they should coincide, since both describe the same contact), and sample $z_i \sim q_i$, $z_j \sim q_j$ via the reparameterization trick for reconstruction. Each sampled latent is decoded both by its \emph{own} sensor's decoder (\emph{self}-reconstruction, e.g.\ $g_i(z_i) \to x_i$) and by the \emph{paired} sensor's decoder (\emph{cross}-reconstruction, e.g.\ $g_j(z_i) \to x_j$): the latent from one finger must reconstruct the other finger's ground-truth signal. At inference, we use the posterior mean $z=\mu_i(x_i)$ as a deterministic latent representation so that the downstream policy receives a stable input.

\paragraph{Training objective.}
Each step jointly optimizes three terms over every sensor pair $(i,j)$:
{\small
\begin{equation}
\label{eq:total_loss}
\mathcal{L}_{\textsc{TactX}} \;=\; \sum_{(i,j)}\!\Big[\, 
\lambda_{\text{recon}}\,\mathcal{L}_{\text{recon}}^{(i,j)} 
\;+\; \alpha(t)\,\mathcal{L}_{\text{align}}^{(i,j)} 
\;+\; \beta(t)\,\mathcal{L}_{\text{KL}}^{(i,j)} \,\Big].
\end{equation}
}

The \emph{reconstruction} term combines the self- and cross-reconstruction flows described above,
{\small
\begin{equation}
\label{eq:recon_loss}
\mathcal{L}_{\text{recon}}^{(i,j)} = 
\underbrace{\|g_i(z_i){-}x_i\|_1 + \|g_j(z_j){-}x_j\|_1}_{\text{self}} 
+ 
\underbrace{\|g_i(z_j){-}x_i\|_1 + \|g_j(z_i){-}x_j\|_1}_{\text{cross}},
\end{equation}
}
where each term is mean-reduced over its target; cross-reconstruction forces shared content through the latent and ties the modalities together.

The \emph{alignment} term is a symmetric NT-Xent loss~\citep{simclr} on L2-normalized posterior means $\tilde\mu_i = \mu_i / \|\mu_i\|_2$ with temperature $\tau{=}0.01$. For a batch of $N$ paired contacts, the two posterior means from the same contact form a positive pair, while the remaining embeddings in the batch serve as negatives.
\begin{equation}
\footnotesize
\label{eq:align_loss}
\mathcal{L}_{\text{align}}^{(i,j)}
=
-\frac{1}{N}\sum_{n=1}^{N}
\log
\frac{
\exp(\tilde\mu_i^{(n)}{\cdot}\tilde\mu_j^{(n)}/\tau)
}{
\sum_{m=1}^{N}
\exp(\tilde\mu_i^{(n)}{\cdot}\tilde\mu_j^{(m)}/\tau)
}.
\end{equation}
The \emph{KL} term regularizes each posterior toward a shared prior $\mathcal{N}(0,I)$, giving all modalities a common target region. We use $\lambda_{\text{recon}}{=}1$, $\alpha(t)$ optionally ramped via a reconstruction-first curriculum, and $\beta(t)$ warmed up from $0$ to $\beta_{\max}{=}0.1$ over the first $30$ epochs. Full schedules are in Appendix~\ref{app:train}.

\paragraph{Pairwise joint training.}
Each trajectory contains only two sensors, so each training step samples from each available pair dataset $\mathcal{D}_{ij}$ and optimizes the resulting paired batches jointly. This exposes every sensor encoder to paired supervision at each step, while the shared latent space and common prior tie the pairwise alignments into a globally consistent representation.

\vspace{-5pt}

\section{Experimental Evaluation}

\textsc{TactX} is designed to align tactile sensors across different sensing modalities while preserving the contact information needed for manipulation. While prior cross-sensor tactile methods largely evaluate transfer within sensors that share a similar sensing substrate, we evaluate \textsc{TactX} across Daimon (vision-based)~\cite{daimon}, eFlesh (magnetic)~\cite{eflesh}, and FlexiTac (resistive)~\cite{flexitac} tactile sensors whose raw observations differ in geometry, dimensionality, and physical sensing mechanisms. We evaluate the learned latent space by testing whether it is sensor-invariant, jointly aligned across all three sensors from pairwise supervision, and still preserves tactile content through object-level prediction and self- and cross-reconstruction. Finally, we test whether the shared latent can serve as a sensor-agnostic tactile interface for Action Chunking with Transformers (ACT) policies~\cite{act}, allowing a policy trained with one tactile sensor to be deployed on another sensor without retraining. Our experiments explore cross-sensor alignment (4.1), three-way alignment from pairwise data (4.2), tactile content preservation (4.3), and zero-shot policy transfer through the shared latent (4.4).

\subsection{How well do different sensors align in a shared representation space?}
\label{sec:q1}

\begin{wrapfigure}{r}{0.5\textwidth}
    \vspace{-8pt}
    \centering
    \includegraphics[width=\linewidth]{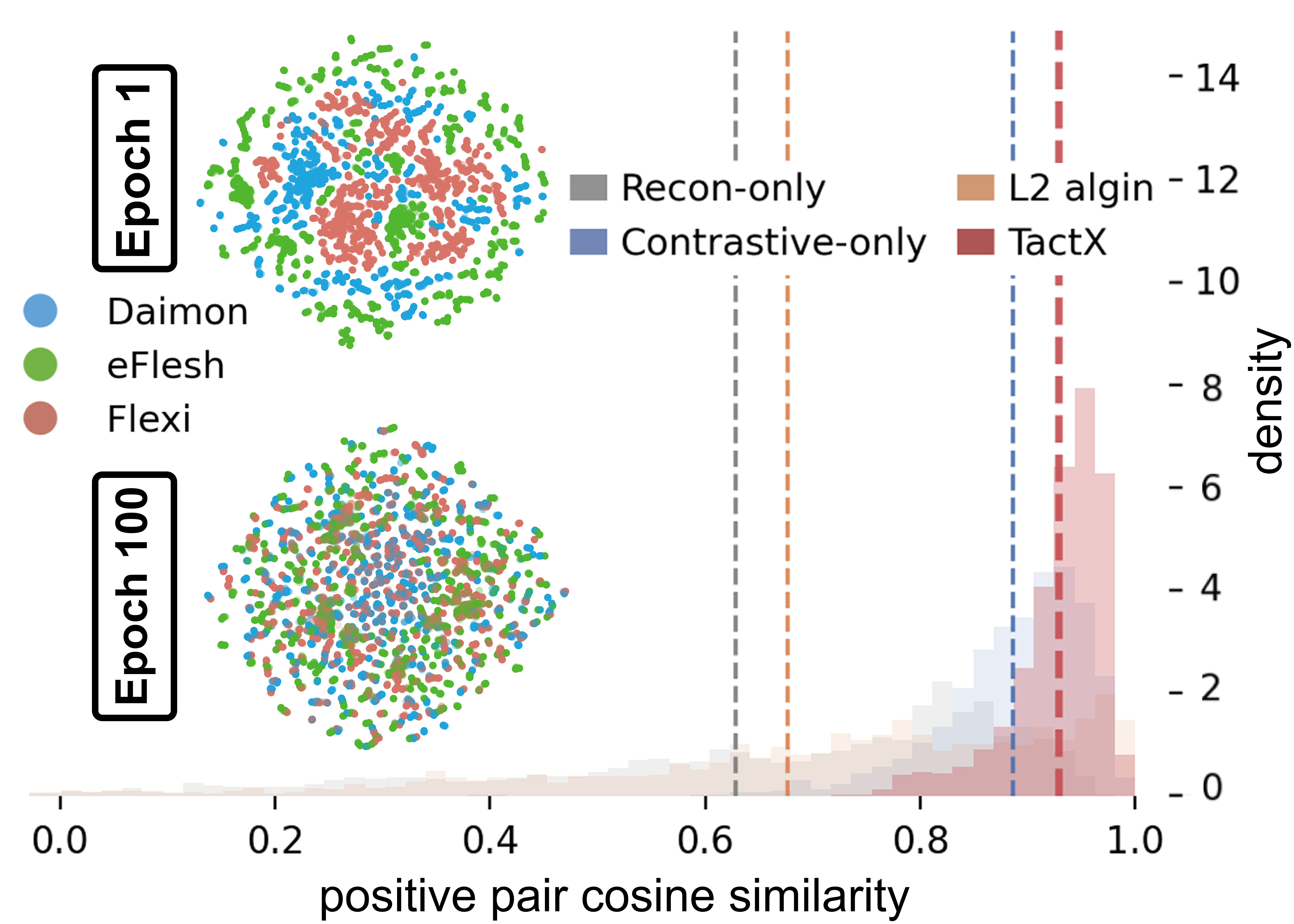}
    \caption{\textbf{Transitive cross-sensor alignment.} Cosine similarity along the Daimon$\rightarrow$eFlesh$\rightarrow$FlexiTac path measures global latent alignment, with dashed lines indicating the mean for each method.}
    \label{fig:align}
    \vspace{-10pt}
\end{wrapfigure}

We first evaluate whether \textsc{TactX} aligns tactile observations from different sensing modalities into a shared latent space. We compare \textsc{TactX} with three objective variants: a reconstruction-only model (using Eq.~\eqref{eq:recon_loss}), a contrastive-only model (using Eq.~\eqref{eq:align_loss}), and an L2-alignment model that replaces the contrastive objective with a direct pull between paired latents. We quantify alignment by computing the cosine similarity between positive contact pairs, and evaluate sensor invariance with a linear sensor-identity probe on the frozen latent.

\begin{figure}[tb]
    \centering
    \includegraphics[width=\linewidth]{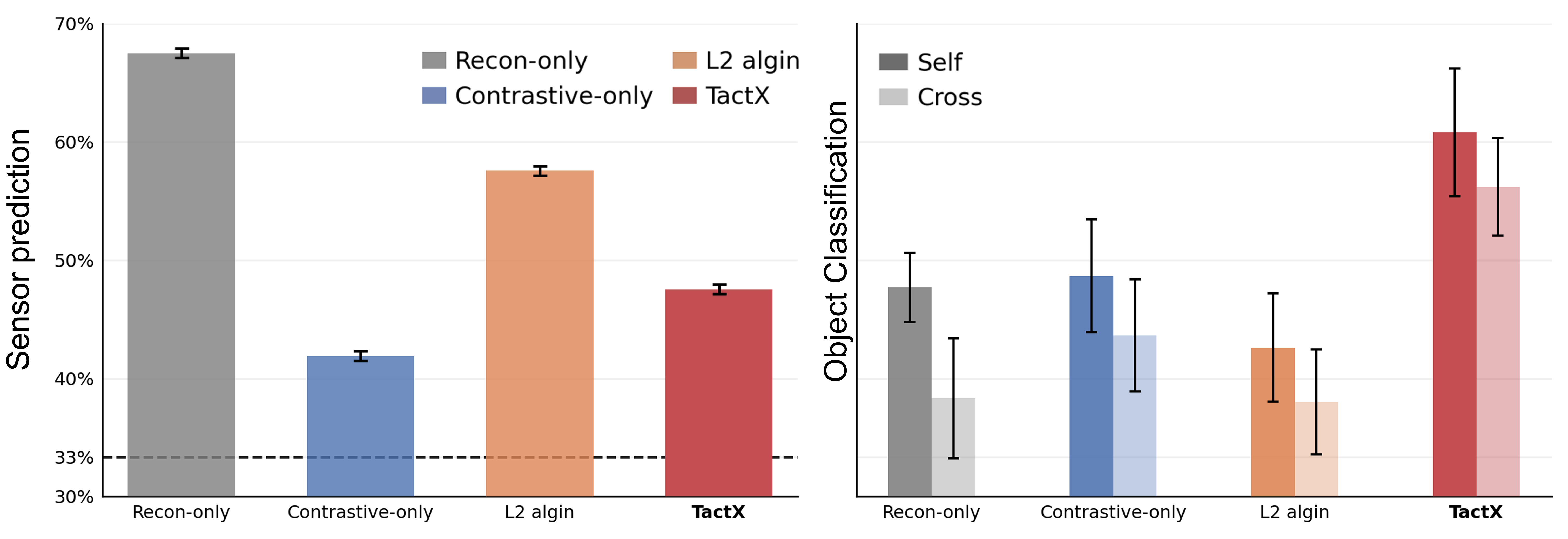}
    \caption{\textbf{Sensor invariance and semantic preservation in the shared latent space.}
    Sensor-prediction accuracy measures whether sensor identity remains recoverable from frozen latents, where lower values closer to the 33.3\% chance level indicate stronger sensor invariance. Object-classification accuracy evaluates whether object-level information is preserved, where ``Self'' denotes training and testing on the same sensor and ``Cross'' denotes training on one sensor and testing on aligned latents from the other sensors.}
    \label{fig:alignment_eval}
\end{figure}

The results in Figure~\ref{fig:alignment_eval} show that \textsc{TactX} consistently aligns paired contacts across all sensor pairs. As expected, contrastive-only training gives high positive-pair similarity, since it is optimized solely to pull paired contacts together. \textsc{TactX} achieves comparable alignment while also retaining the reconstruction objective, which is important for preserving tactile content as evaluated in Section~\ref{sec:q3}.

The t-SNE visualization shows the same trend qualitatively. Before training, samples from different sensors occupy separate regions, reflecting the gap between their raw sensing modalities. After training, the three sensor domains become substantially more mixed. The sensor-prediction probe further supports this observation. We train this probe as a linear classifier on the learned latent space with the encoders frozen, so only the classifier is trained to predict sensor identity. Because lower sensor-prediction accuracy indicates stronger sensor invariance, the ideal representation should approach the 33.3\% chance level. \textsc{TactX} reduces sensor prediction accuracy from 67.5\% for reconstruction-only training to 47.5\%, the closest to chance among the reconstruction-based variants. These results indicate that \textsc{TactX} learns a shared latent space in which heterogeneous tactile sensors are well aligned.

\subsection{Can pairwise data align multiple sensors?}
\label{sec:q2}

We next ask whether pairwise supervision is sufficient to produce a globally consistent tactile space of multiple sensors. Our data are collected from two sensors at a time, so the model never observes optical, magnetic, and resistive tactile readings from the same contact simultaneously. This makes three-way alignment nontrivial: the model must align each observed pair while placing all three modalities into a single shared coordinate system.

We evaluate this property using transitive alignment between Daimon and FlexiTac through eFlesh. We denote Daimon, eFlesh, and FlexiTac as D, E, and F, respectively. Each positive pair contains two observations of the same contact, one from each sensor. The model is trained on D--E and E--F positive pairs, but does not observe D--F contacts jointly in this test; therefore, successful D--F alignment requires consistency through the shared E bridge. For each E--F pair, we find the nearest E latent in the D--E set, take its paired D latent, and measure the cosine similarity between this matched D latent and the F latent. Figure~\ref{fig:align} shows that \textsc{TactX} achieves the strongest transitive alignment, increasing the D--F cosine from 0.626 with reconstruction-only training and 0.679 with L2-alignment to 0.928. This indicates that the model does not simply learn independent pairwise mappings, but instead induces a globally consistent shared latent space across multiple tactile modalities from pairwise supervision.

\subsection{Does the shared latent preserve critical tactile information?}
\label{sec:q3}

We next evaluate whether the shared latent space preserves tactile content, rather than only removing sensor identity. This is important because a fully sensor-invariant representation is useful for manipulation only if it still retains information about the underlying contact geometry. We examine this property through reconstruction from the latent space. For each target sample in the validation set, we decode the sensor's own latent for self-reconstruction. For cross-reconstruction, we do not use the paired observation directly; instead, we retrieve the nearest latent from the other sensor in the validation set and decode it into the target modality.

\begin{figure}[tb]
    \centering
    \includegraphics[width=\linewidth]{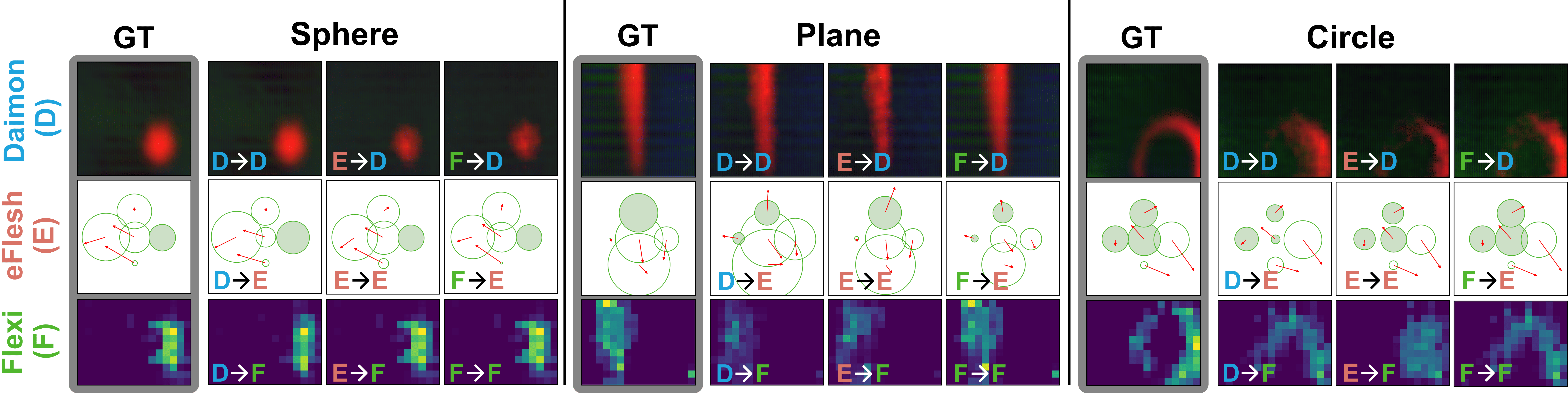}
    \caption{\textbf{Self- and cross-reconstruction from the shared latent.}
    We visualize representative validation contacts from sphere, plane, and circle indentors. For each sensor, the first column is the ground-truth observation, the diagonal entries are self-reconstructions, and the off-diagonal entries are cross-reconstructions decoded from the nearest latent representations of the other sensors in the validation set.}
    \label{fig:reconstruction}
\end{figure}

The results in Figure~\ref{fig:reconstruction} show that \textsc{TactX} preserves the dominant contact patterns across all three sensing modalities. Self-reconstructions recover the original contact structure, and cross-reconstructions from nearest aligned latents remain visually consistent with the corresponding ground truth. This indicates that the shared latent does not merely discard sensor-specific information, but retains object- and contact-level structure that can be decoded across modalities.

We further evaluate tactile content preservation by training object classifiers on frozen latent features and testing them on unseen contact points from 10 object classes. The results in Figure~\ref{fig:alignment_eval} show that \textsc{TactX} achieves the highest accuracy in both settings, reaching 60.8\% for self-sensor evaluation and 56.2\% for cross-sensor evaluation. This indicates that the shared latent preserves object-level tactile structure even when evaluated across sensors.


\subsection{Can robot policies transfer across sensors through the shared latent?}
\label{sec:q4}

We finally evaluate whether the shared latent space can support tactile-conditioned policy learning across physically different sensors. We consider four contact-rich manipulation tasks: pick-and-place, plug insertion, board wiping, and object reorientation. For pick-and-place, we additionally evaluate an out-of-distribution color setting, where policies trained on black objects are tested on white objects with the same geometry. For each task, we train an ACT policy~\cite{act} using demonstrations from one tactile sensor and evaluate it under a different tactile sensor.

\begin{wraptable}{r}{0.48\textwidth}
\vspace{-10pt}
\centering
\caption{
\textbf{Average same-sensor policy performance.}
Results are averaged over three same-sensor train-test evaluations.
}
\vspace{-2pt}
\label{tab:same_sensor_task_average_summary}
\vspace{3pt}
\footnotesize
\setlength{\tabcolsep}{3.5pt}
\renewcommand{\arraystretch}{1.08}
\begin{tabular}{lccc}
\toprule
\textbf{Task}
& \textbf{Vision}
& \textbf{+ Tactile GT}
& \textbf{+ \textsc{TactX}} \\
\midrule
P\&P
& 8.33
& 9.33
& \textbf{10.00} \\

P\&P (OOD)
& 6.67
& \textbf{8.00}
& 7.33 \\

Insertion
& 4.00
& \textbf{7.33}
& 6.00 \\

Wiping
& 4.33
& \textbf{8.33}
& 7.33 \\

Reorientation
& 8.00
& 9.33
& \textbf{9.67} \\
\bottomrule
\end{tabular}
\vspace{-13pt}
\end{wraptable}

Table~\ref{tab:same_sensor_task_average_summary} summarizes the policy results. In the in-domain setting, tactile input improves over vision-only policies, confirming that contact information is useful for these tasks.
Raw tactile observations provide a strong same-sensor 
performance, and we treat this as an oracle upper bound: the goal of cross-sensor transfer 
is to \emph{approach} this value. \textsc{TactX} retains most of this 
benefit using a shared latent representation in place of sensor-specific raw inputs.

The main result is cross-sensor transfer, shown in Table~\ref{tab:cross_sensor_policy_results}. Vision-only policies provide a sensor-independent baseline but degrade on contact-rich or visually shifted settings. Binary contact transfer tests whether a minimal contact/no-contact signal is sufficient, but it removes spatial and geometric contact information. In contrast, \textsc{TactX} achieves the best cross-sensor performance on most tasks and sensor-pairs when transferring between sensors zero-shot. Averaged over all transfer directions and tasks, \textsc{TactX} improves the success rate from 27.5\% to 45.9\% over vision-only transfer.

\newcommand{\g}{\cellcolor{gray!15}}
\begin{table}[t]
\centering
\caption{
\textbf{Cross-sensor policy transfer across all tasks.}
Each entry reports the number of successes out of 10 trials, shown as mean $\pm$ std over 3 runs.
\textbf{Bold} indicates the best performance for each source-deploy combination and task.
}
\label{tab:cross_sensor_policy_results}
\vspace{5pt}
\begingroup
\footnotesize
\setlength{\tabcolsep}{3pt}
\renewcommand{\arraystretch}{1.05}

\begin{tabular}{l l l c c c c c}
\hline
\textbf{Method}
& \textbf{Source}
& \textbf{Deploy}
& \textbf{P\&P}
& \textbf{P\&P (OOD)}
& \textbf{Insertion}
& \textbf{Wiping}
& \textbf{Reorient} \\
\hline

\multirow{6}{*}{\shortstack{Vision\\Transfer}}
& \multirow{2}{*}{\Daimon}
& \eFlesh
& 5.3 $\pm$ 0.9
& \textbf{1.7 $\pm$ 0.9}
& 2.7 $\pm$ 0.5
& 3.0 $\pm$ 0.0
& 0.3 $\pm$ 0.5 \\
&
& \FlexiTac
& 5.3 $\pm$ 0.5
& 1.3 $\pm$ 0.5
& 1.3 $\pm$ 0.9
& 1.3 $\pm$ 0.9
& 0.7 $\pm$ 0.9 \\

& \g
& \g\Daimon
& \g 7.7 $\pm$ 0.5
& \g 0.0 $\pm$ 0.0
& \g 3.7 $\pm$ 0.5
& \g 0.0 $\pm$ 0.0
& \g \textbf{7.0 $\pm$ 0.0} \\
& \g\multirow{-2}{*}{\eFlesh}
& \g\FlexiTac
& \g 1.3 $\pm$ 0.5
& \g 0.7 $\pm$ 0.5
& \g 0.3 $\pm$ 0.5
& \g 0.3 $\pm$ 0.5
& \g 1.3 $\pm$ 0.5 \\

& \multirow{2}{*}{\FlexiTac}
& \Daimon
& 7.0 $\pm$ 0.8
& 6.0 $\pm$ 0.0
& 3.3 $\pm$ 0.5
& 0.7 $\pm$ 0.5
& 7.3 $\pm$ 0.9 \\
&
& \eFlesh
& 6.7 $\pm$ 0.5
& 1.0 $\pm$ 0.0
& 5.0 $\pm$ 0.0
& 0.3 $\pm$ 0.5
& 0.0 $\pm$ 0.0 \\
\hline

\multirow{6}{*}{\shortstack{Binary\\Contact\\Transfer}}
& \multirow{2}{*}{\Daimon}
& \eFlesh
& 4.0 $\pm$ 0.0
& 1.3 $\pm$ 0.9
& 2.7 $\pm$ 2.1
& 2.0 $\pm$ 2.8
& 1.7 $\pm$ 1.7 \\
&
& \FlexiTac
& 3.3 $\pm$ 1.2
& \textbf{2.7 $\pm$ 0.5}
& \textbf{2.0 $\pm$ 0.8}
& \textbf{2.0 $\pm$ 1.4}
& 4.0 $\pm$ 1.6 \\

& \g
& \g\Daimon
& \g 3.3 $\pm$ 0.9
& \g 0.0 $\pm$ 0.0
& \g 0.3 $\pm$ 0.5
& \g 1.0 $\pm$ 1.4
& \g 2.3 $\pm$ 2.1 \\
& \g\multirow{-2}{*}{\eFlesh}
& \g\FlexiTac
& \g \textbf{4.0 $\pm$ 1.6}
& \g \textbf{1.3 $\pm$ 1.2}
& \g 0.7 $\pm$ 0.9
& \g 0.3 $\pm$ 0.5
& \g \textbf{6.0 $\pm$ 2.2} \\

& \multirow{2}{*}{\FlexiTac}
& \Daimon
& 3.3 $\pm$ 1.7
& 2.7 $\pm$ 1.7
& 6.7 $\pm$ 0.5
& 1.3 $\pm$ 1.9
& 7.3 $\pm$ 2.4 \\
&
& \eFlesh
& 1.0 $\pm$ 0.8
& 2.7 $\pm$ 1.9
& 0.3 $\pm$ 0.5
& 1.0 $\pm$ 0.8
& 2.7 $\pm$ 0.9 \\
\hline

\multirow{6}{*}{\shortstack{\textbf{TactX}\\\textbf{Transfer}\\\textbf{(Ours)}}}
& \multirow{2}{*}{\Daimon}
& \eFlesh
& \textbf{8.3 $\pm$ 0.5}
& 1.0 $\pm$ 0.0
& \textbf{4.0 $\pm$ 0.8}
& \textbf{4.0 $\pm$ 0.0}
& \textbf{3.7 $\pm$ 0.9} \\
&
& \FlexiTac
& \textbf{5.3 $\pm$ 1.2}
& 2.0 $\pm$ 0.8
& 1.3 $\pm$ 1.9
& 0.3 $\pm$ 0.5
& \textbf{6.7 $\pm$ 0.9} \\

& \g
& \g\Daimon
& \g \textbf{9.0 $\pm$ 1.4}
& \g \textbf{0.7 $\pm$ 0.5}
& \g \textbf{6.0 $\pm$ 1.6}
& \g \textbf{6.0 $\pm$ 0.8}
& \g 5.0 $\pm$ 1.4 \\
& \g\multirow{-2}{*}{\eFlesh}
& \g\FlexiTac
& \g \textbf{1.3 $\pm$ 0.9}
& \g 0.0 $\pm$ 0.0
& \g \textbf{3.7 $\pm$ 1.2}
& \g \textbf{1.3 $\pm$ 0.5}
& \g 5.0 $\pm$ 0.8 \\

& \multirow{2}{*}{\FlexiTac}
& \Daimon
& \textbf{8.0 $\pm$ 1.4}
& \textbf{8.3 $\pm$ 1.2}
& \textbf{8.3 $\pm$ 0.9}
& \textbf{6.3 $\pm$ 0.9}
& \textbf{7.7 $\pm$ 1.2} \\
&
& \eFlesh
& \textbf{6.7 $\pm$ 0.9}
& \textbf{3.0 $\pm$ 0.8}
& \textbf{4.7 $\pm$ 0.5}
& \textbf{5.7 $\pm$ 0.5}
& \textbf{4.3 $\pm$ 0.5} \\
\hline

\end{tabular}

\endgroup
\end{table}

\begin{figure*}[t]
    \centering
    \begin{subfigure}[t]{0.24\textwidth}
        \centering
        \includegraphics[width=\linewidth]{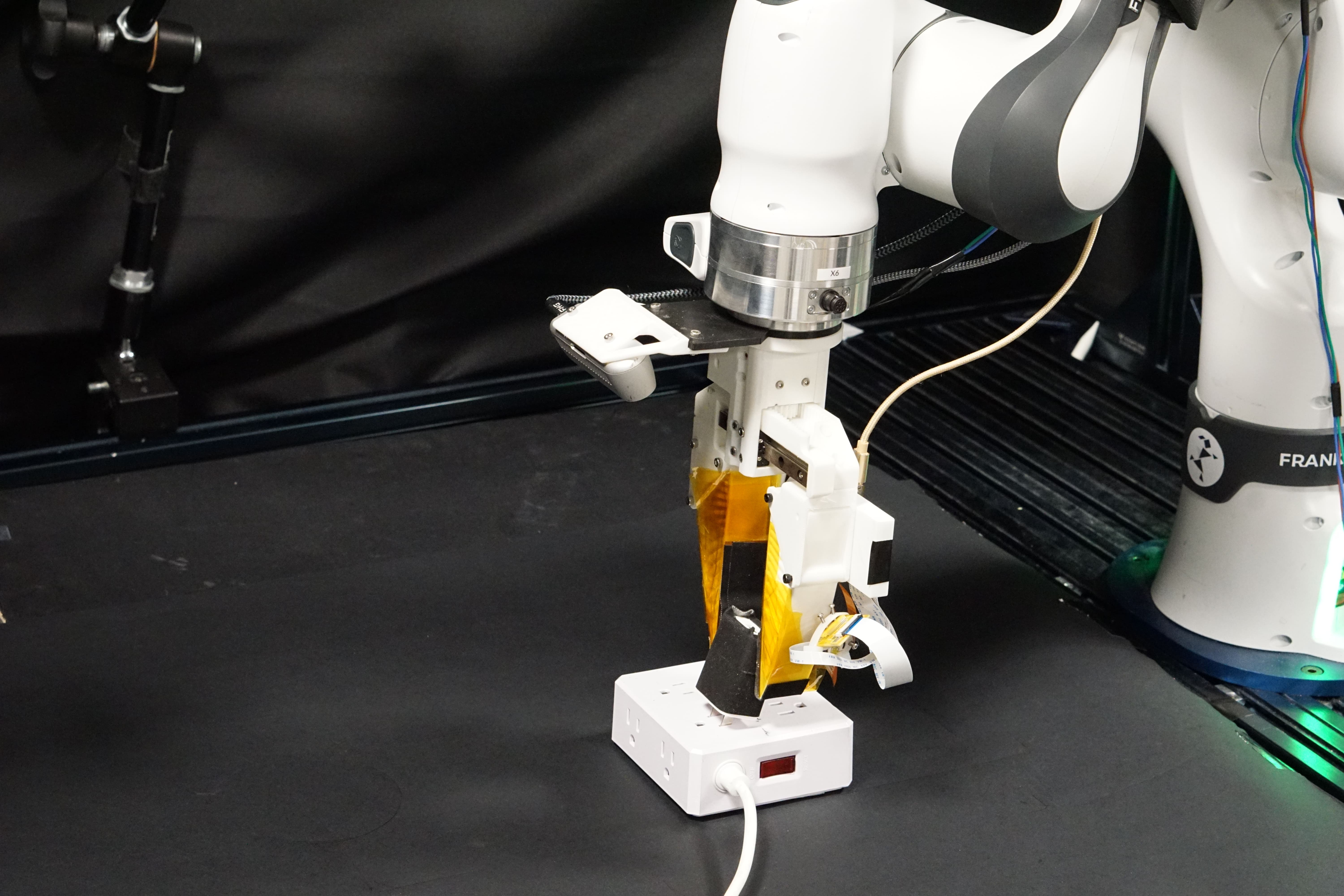}
        \caption{Plug insertion}
    \end{subfigure}
    \hfill
    \begin{subfigure}[t]{0.24\textwidth}
        \centering
        \includegraphics[width=\linewidth]{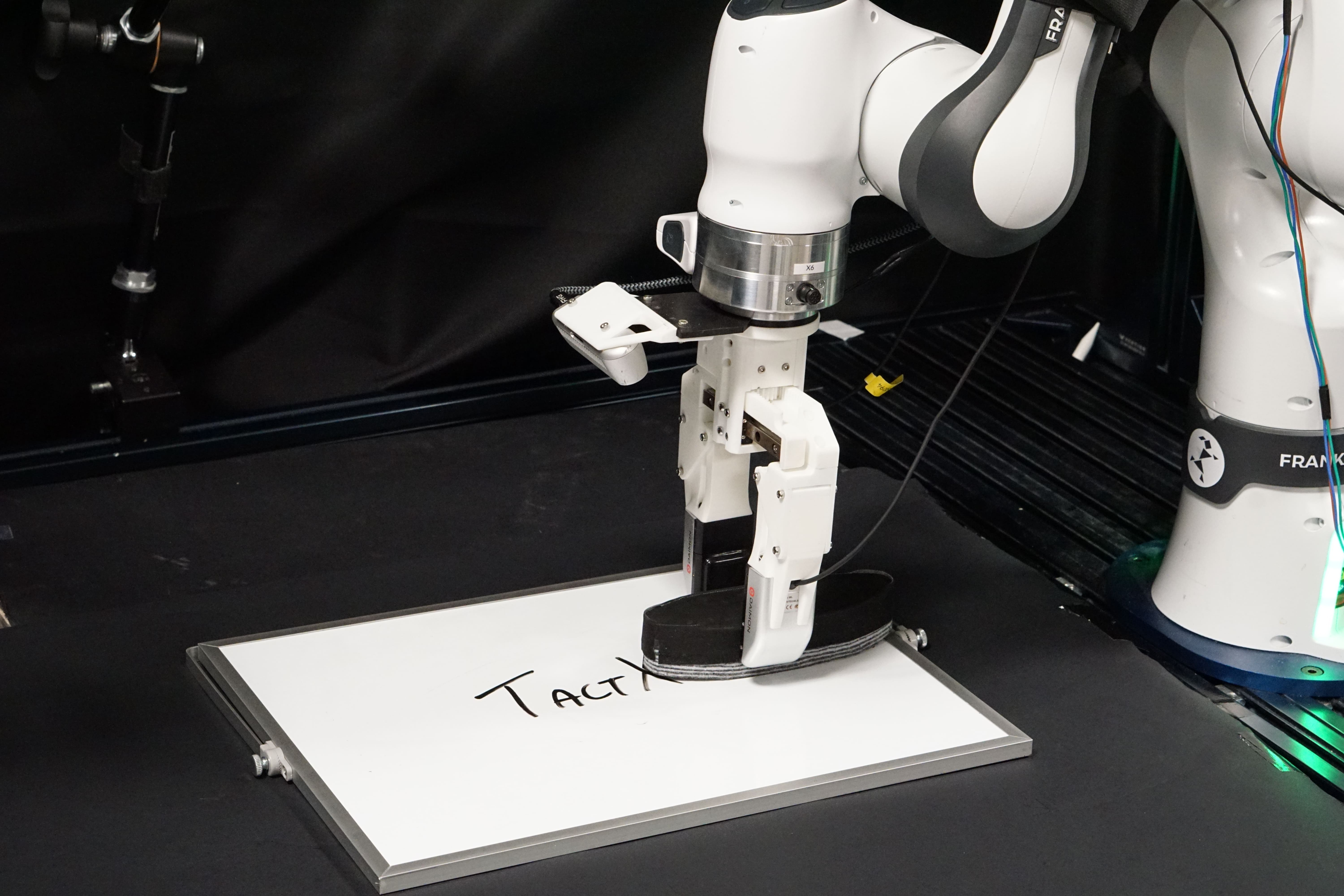}
        \caption{Board wiping}
    \end{subfigure}
    \hfill
    \begin{subfigure}[t]{0.24\textwidth}
        \centering
        \includegraphics[width=\linewidth]{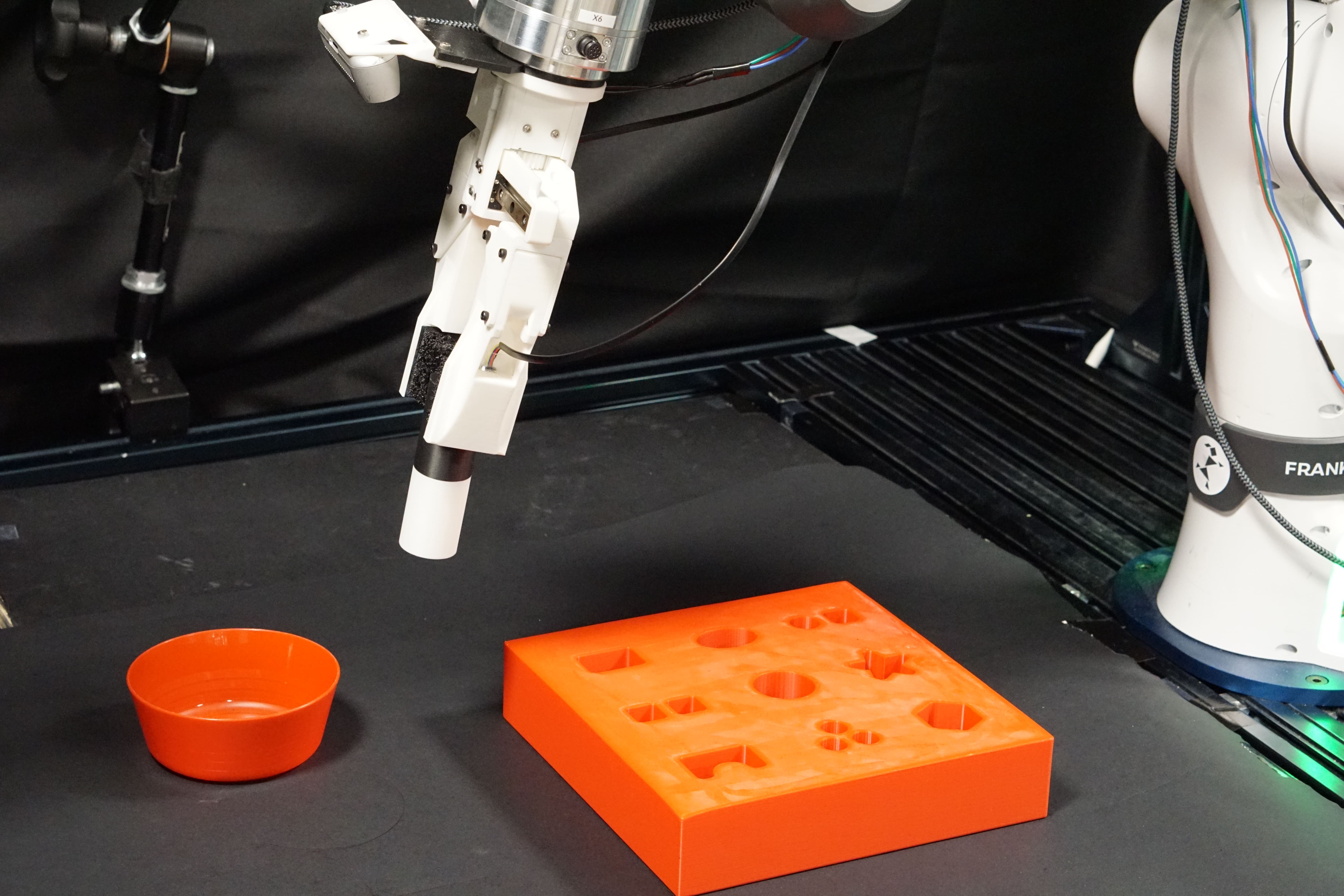}
        \caption{Pick-and-place}
    \end{subfigure}
    \hfill
    \begin{subfigure}[t]{0.24\textwidth}
        \centering
        \includegraphics[width=\linewidth]{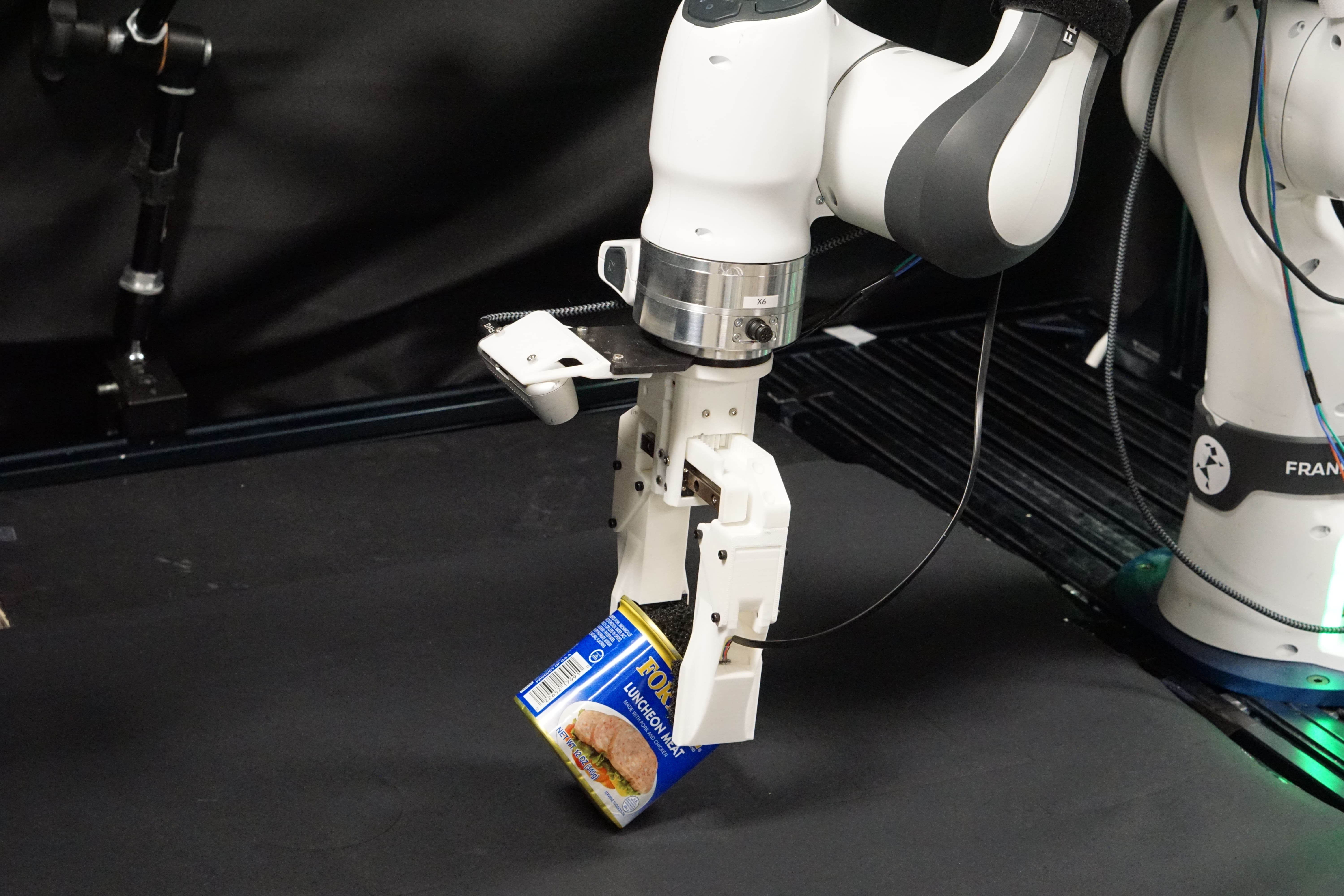}
        \caption{Object reorientation}
    \end{subfigure}

    \caption{\textbf{Downstream manipulation tasks.}
    We evaluate zero-shot tactile policy transfer on four contact-rich tasks: plug insertion, board wiping, pick-and-place, and object reorientation.}
    \label{fig:tasks}
\end{figure*}

Contact-rich tasks such as board wiping and object reorientation show the largest 
improvements over the vision-only baseline. Binary contact transfer provides only limited 
benefit in these settings, suggesting that the shared latent captures richer contact 
geometry than a simple contact/no-contact signal and recovers failures that vision alone 
cannot resolve. A further limitation of binary contact transfer is its sensitivity to the 
contact threshold: we use three separate sensor-specific thresholds that are held fixed 
across all tasks (Appendix~\ref{app:policy_contact}), and this threshold mismatch between tasks leads to higher 
variance and inconsistent results across task conditions.

We also observe that transfer is not symmetric across sensors. The weakest direction is eFlesh to FlexiTac, where all methods perform poorly. This suggests that a policy trained with the lower-dimensional magnetic signal may not learn to use the finer spatial structure available from the resistive sensor at deployment. The reverse is stronger, indicating that policies trained with a richer tactile representation perform more gracefully when deployed with a lower-bandwidth sensor. Overall, these results suggest that \textsc{TactX} does not simply improve latent metrics, but provides a practical shared representation for zero-shot tactile policy transfer across heterogeneous sensors.

\section{Conclusion}
\label{sec:conclusion}

We introduced \textsc{TactX}, a framework for learning sensor-agnostic tactile representations for contact-rich manipulation. \textsc{TactX} aligns heterogeneous tactile observations from vision-based, magnetic, and resistive sensors into a shared latent space using paired contact data, contrastive alignment, and reconstruction. By addressing sensors with fundamentally different tactile transduction modalities, \textsc{TactX} goes beyond transfer among visually similar tactile sensors and enables a common representation for structurally diverse touch signals. This shared representation enables zero-shot cross-sensor policy transfer, allowing policies trained with one tactile sensor to be deployed with a physically distinct sensor without retraining. Across multiple manipulation tasks, \textsc{TactX} improves zero-shot cross-sensor policy transfer over vision-only baselines, demonstrating that aligned tactile representations can help policies generalize across heterogeneous tactile hardware. These results suggest a scalable path toward shared tactile representations that reduce the need to collect new demonstrations and retrain policies for each new sensor.
\section{Limitations}
\label{sec:limitations}

While our results demonstrate the feasibility of zero-shot cross-sensor tactile policy transfer, \textsc{TactX}
still has several limitations. First, our method relies on paired contact data to align heterogeneous
tactile sensors, which requires different sensors to observe corresponding contact events under comparable
object poses, contact locations, and interaction conditions. This assumption may be more difficult to
satisfy for asymmetric or geometrically complex objects, where two sensors can produce different tactile
readings due to differences in placement or local contact geometry rather than sensing modality alone. Second, our current data is collected primarily from quasi-static gripping
interactions, which provide clean alignment supervision but do not fully capture the dynamic contact
variations that arise during manipulation. For example, although \textsc{TactX} transfers effectively on board
wiping overall, failures can occur under large shear changes or sustained sliding contact. Future work will
address these limitations by exploring weaker forms of supervision, such as weakly paired, unpaired, or
self-supervised alignment, and by extending data collection to dynamic tactile interactions such as sliding,
pushing, and other contact-rich motions.

\newpage
\bibliography{ref}
\newpage
\appendix

\section{Data Collection Details}
\label{app:data}

\paragraph{Sensors.}
To prevent too much visual change the eFlesh housing is 3D-printed in black TPU and the FlexiTac surface is covered with black anti-slip tape, matching Daimon's black elastomer; the three active sensing areas are roughly commensurate so a contact is captured by every sensor.

\begin{figure}[h]
    \centering
    \begin{subfigure}[t]{0.31\textwidth}
        \centering
        \includegraphics[width=\linewidth, height=0.18\textheight, keepaspectratio]{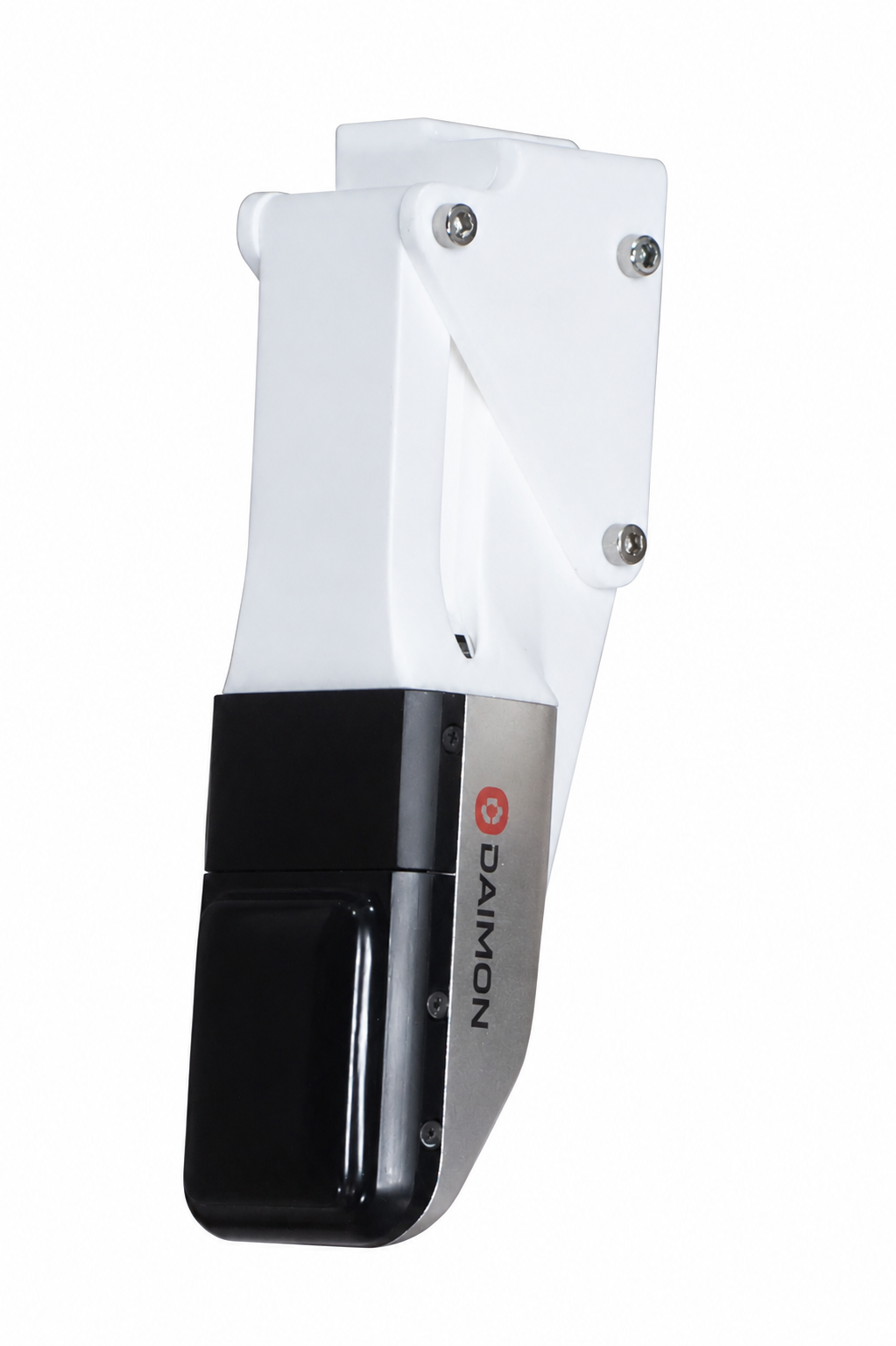}
        \caption{Daimon (vision-based)}
        \label{fig:app_sensor_d}
    \end{subfigure}\hfill
    \begin{subfigure}[t]{0.31\textwidth}
        \centering
        \includegraphics[width=\linewidth, height=0.18\textheight, keepaspectratio]{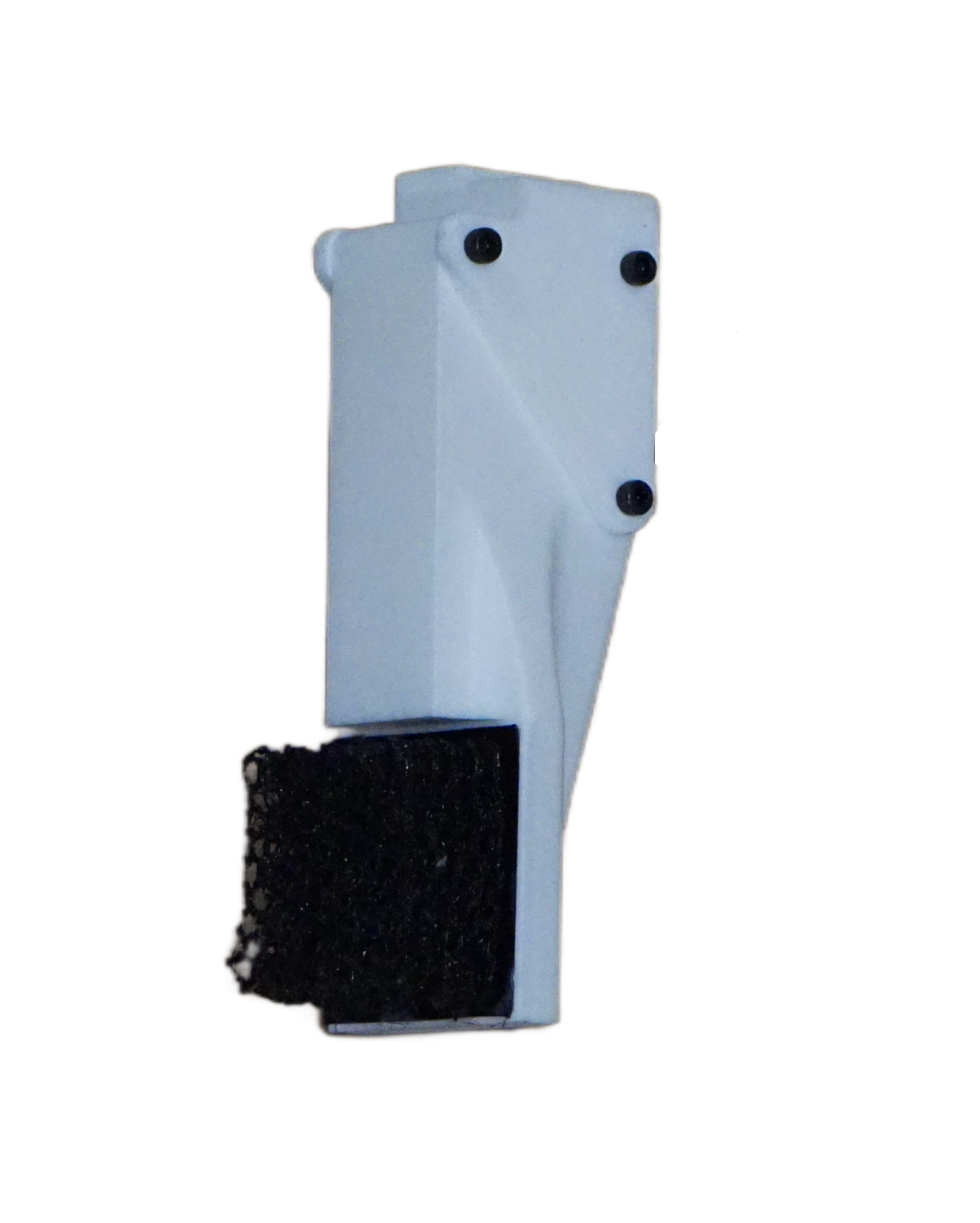}
        \caption{eFlesh (magnetic)}
        \label{fig:app_sensor_e}
    \end{subfigure}\hfill
    \begin{subfigure}[t]{0.31\textwidth}
        \centering
        \includegraphics[width=\linewidth, height=0.18\textheight, keepaspectratio]{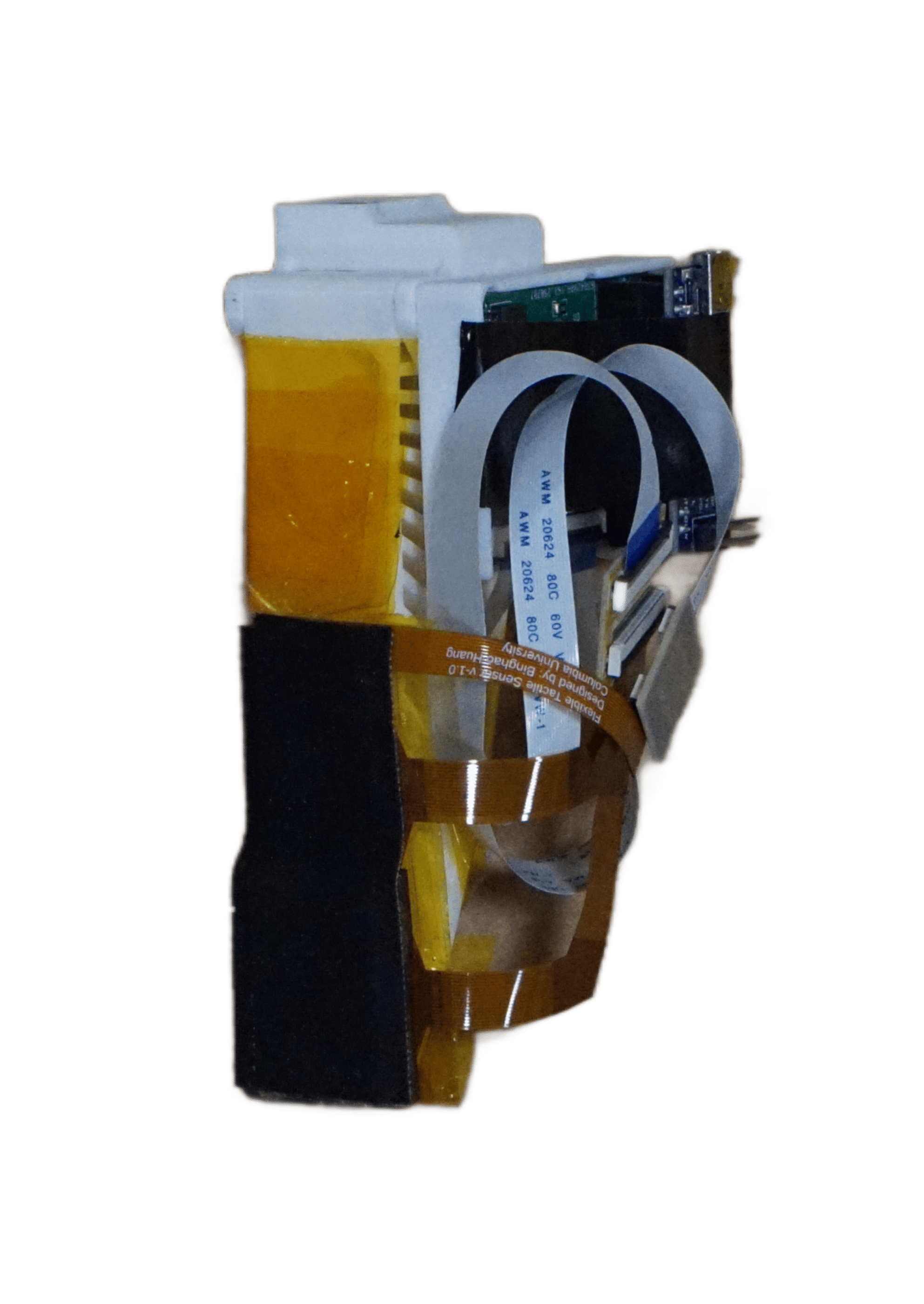}
        \caption{FlexiTac (resistive)}
        \label{fig:app_sensor_f}
    \end{subfigure}
    \caption{The three tactile sensors used in \textsc{TactX}, each spanning a different transduction modality. All three are visually matched (black TPU/tape/elastomer) to remove cosmetic shortcuts and have roughly commensurate active sensing areas.}
    \label{fig:app_sensors}
\end{figure}

\paragraph{Mounting and pairing.}
Two sensors are mounted on opposing fingers of a Franka parallel-jaw gripper; the third is swapped in for separate runs. We cover all $\binom{3}{2} \times 2 = 6$ configurations: each unordered pair is recorded twice with sensors swapped between the left and right fingers, doubling the effective pair-dataset size and removing left/right asymmetry. This yields three pair-datasets $\mathcal{D}_{DE}, \mathcal{D}_{EF}, \mathcal{D}_{FD}$. Because the sensors sit at structurally different positions on their respective housings, we apply a one-time $180^\circ$ rotation to one sensor of each pair at load time so contact regions align across paired observations.

\paragraph{Objects and protocol.}
\begin{wrapfigure}{r}{0.38\columnwidth}
    \centering
    \vspace{-10pt}
    \includegraphics[width=\linewidth]{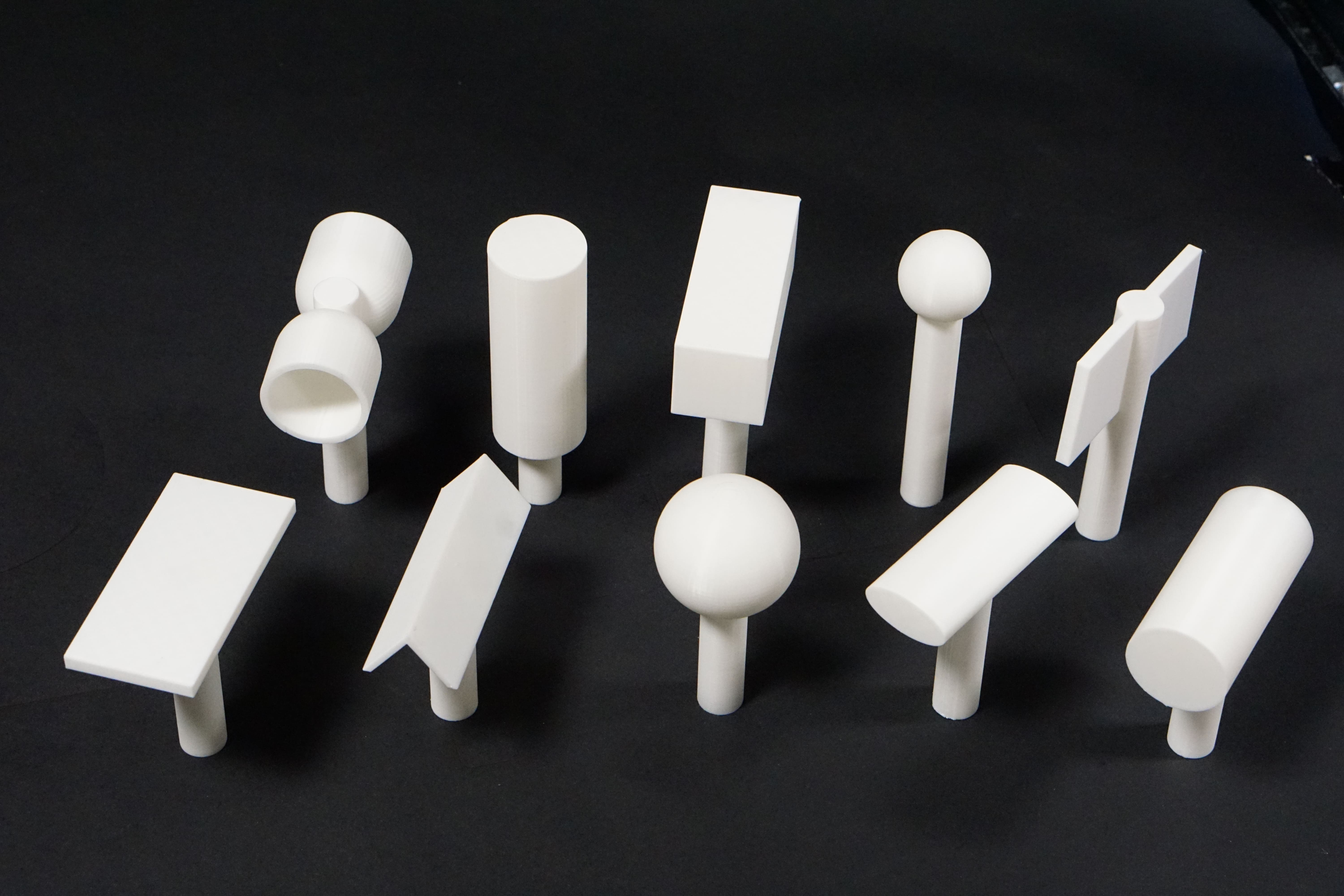}
    \caption{The 10 3D-printed pretraining objects used for paired data collection, spanning point, edge, and area contact geometries.}
    \label{fig:app_objects}
    \vspace{-10pt}
\end{wrapfigure}

Pretraining data uses 10 3D-printed objects (Fig.~\ref{fig:app_objects}) spanning point contacts (small ball), edge contacts (line, triangle, circle circumference), and area contacts (large ball, ellipse), sized to fit within the shared sensing area. Each grasp follows a scripted approach--contact--stable-grasp--release--withdraw protocol; automation ensures consistent contact conditions across the many repetitions needed for paired supervision. All three sensors are sampled at a matched rate and paired frames are aligned by nearest timestamp.

\paragraph{Contact gating.}
At preprocessing, frames are labeled \emph{contact} if Daimon depth abs-mean exceeds $0.003$ or FlexiTac pressure grid mean exceeds $0.01$; otherwise no-contact.

\begin{table}[h]
\centering
\caption{Pretraining data composition.}
\label{tab:app_data}
\begin{tabular}{ll}
\toprule
Objects (10) & circle, cylinder, cylinder\_vertical, ellipse, large\_ball, \\
             & plane, plane\_horizontal, small\_ball, square, triangle \\
Pair-datasets & $\mathcal{D}_{DE},\ \mathcal{D}_{EF},\ \mathcal{D}_{FD}$ (6 mounting configs, L/R-swapped) \\
Trajectories & 2,670 total ($\sim$442--448 / config, across 10 objects) \\
Frames       & 145k total ($\sim$20k--31k / config, $\sim$54 / trajectory) \\
Train/val split & episode-level, $\sim$20\% val, sampled across the recording order \\
\bottomrule
\end{tabular}
\end{table}

\section{\textsc{TactX} Architecture}
\label{app:arch}

Each encoder maps its native signal to a 512-D feature, then a shared-form projection head ($512\!\to\!512\!\to\!2d$, Linear--ReLU--Linear) outputs $(\mu,\log\sigma^2)$ over the $d{=}16$ latent. At training $z=\mu+\sigma\odot\epsilon,\ \epsilon\sim\mathcal{N}(0,I)$; at inference $z=\mu$. Decoders are sensor-specific (no shared decoder, since output spaces differ too much) and each is used for both self- and cross-reconstruction. All modules train from scratch.

\begin{table}[h]
\centering
\caption{Per-sensor encoder / decoder architectures. Shared latent $d{=}16$.}
\label{tab:app_arch}
\resizebox{\textwidth}{!}{%
\begin{tabular}{lllll}
\toprule
Sensor & Modality & Native input & Encoder backbone & Decoder \\
\midrule
Daimon (D)   & vision-based & $224{\times}224{\times}3$ (depth+shear) & ResNet-18 $\to$ 512 & linear $\to$ transposed conv \\
eFlesh (E)   & magnetic     & $15$-D field vector & MLP $[64,128,256]\to512$ & reversed MLP \\
FlexiTac (F) & resistive    & $12{\times}16$ pressure grid & residual CNN $\to$ 512 & mirrored conv \\
\bottomrule
\end{tabular}}
\end{table}

\section{\textsc{TactX} Training}
\label{app:train}

\begin{table}[h]
\centering
\caption{\textsc{TactX} representation-learning hyperparameters.}
\label{tab:app_hparams}
\begin{tabular}{ll}
\toprule
Latent dim $d$ & 16 \\
Batch size & 64 \\
Learning rate & $1\times10^{-4}$ (Adam, weight decay $1\times10^{-4}$) \\
Epochs & 300 \\
Seed & 42 \\
$\lambda_{\text{recon}}$ & 1.0 \\
KL weight $\beta$ & $0\to0.1$ linear warmup over 30 epochs \\
InfoNCE temperature $\tau$ & 0.01 (NT-Xent variant: 0.03) \\
Alignment weight $\lambda_{\text{align}}$ & $0\to1$ warmup (start 0) \\
\bottomrule
\end{tabular}
\end{table}

\section{Downstream Policy Details}
\label{app:downstream_policy}

\paragraph{Overview.}
All manipulation policies are Action Chunking Transformers (ACT) with a DETR-style CVAE decoder.
Demonstrations are collected via GELLO teleoperation ($\sim$50 episodes per task, $\sim$10k frames each).
Unless noted, training uses learning rate $1\times10^{-5}$, batch size 8, and 50{,}000 training steps with action-chunk length 64.
Robot state and actions are padded to 128 dimensions; only the first 8 joint/gripper dimensions are used at deployment.
Two RGB cameras (cam0 wrist, cam1 third-person or side-view (for board wiping)) and two tactile fingers (tac0, tac1) are always logged; how tactile data enter ACT depends on the variant below.

\paragraph{Shared ACT backbone.}
Visual observations are encoded with an ImageNet-pretrained ResNet-18 and ACT\_linear projection into hidden dimension 512.
Images are resized to $224\times308$ ($16{\times}14 \times 22{\times}14$ patches) and ImageNet-normalized.
The transformer uses 4 encoder layers, 7 decoder layers, 8 attention heads, hidden dimension 512, and feed-forward width 3200.
The training objective is an L1 loss on predicted action chunks plus $\lambda_{\mathrm{KL}}$ times the KL divergence on ACT's internal CVAE latent $z \in \mathbb{R}^{32}$, which encodes action-sequence diversity and is set to zero at inference.
This CVAE latent is distinct from any tactile representation (raw sensor data, binary contact, or \textsc{TactX} latents).

\begin{table}[t]
\centering
\caption{Shared ACT hyperparameters (all tactile variants).}
\label{tab:app_act_shared}
\begin{tabular}{ll}
\toprule
Policy class & ACT (DETR-style action chunking) \\
State / action dim & 128 / 128 (padded; first 8 used) \\
Enc.\ / dec.\ layers & 4 / 7 \quad nheads 8 \quad hidden dim 512 \quad FFN 3200 \\
Image backbone & ResNet-18 (ImageNet-pretrained), ACT\_linear features \\
Cameras & cam0 (wrist), cam1 (third-person or side-view) \\
Tactile fingers & tac0, tac1 \\
Chunk size / queries & 64 \\
Learning rate & $1\times10^{-5}$ \\
Batch size / steps & 8 / 50{,}000 \\
Demonstrations & $\sim$50 / task via GELLO teleoperation~\citep{gello} \\
\bottomrule
\end{tabular}
\end{table}

\subsection{Raw tactile policies}
\label{app:policy_raw}

Raw policies consume sensor-native measurements per finger with sensor-specific encoders but otherwise share the hyperparameters in Table~\ref{tab:app_act_shared}.
We set $\lambda_{\mathrm{KL}} = 10$ (ACT default) for all raw runs.

\begin{itemize}
  \item \textbf{Daimon (D, vision-style).}
  Modality image: each finger provides a $240\times320\times3$ composite visualization (depth, deformation, and shear panels).
  A second ResNet-18 tactile backbone ($\ell_{\mathrm{tactile}} = 10^{-4}$) encodes these images; projected features are concatenated with camera feature maps along the spatial width axis, matching the layout of additional camera views.
  The camera backbone learning rate is $10^{-5}$.

  \item \textbf{FlexiTac (F, array).}
  Modality array: a $12\times16 = 192$-dimensional resistive grid per finger ( $\sim$30\,Hz).
  Each finger is mapped by an MLP adapter $192 \!\to\! 64 \!\to\! 128 \!\to\! 512$ to one tactile token prepended to the transformer (alongside proprioception and the CVAE token).
  The camera backbone learning rate is $10^{-5}$.

  \item \textbf{eFlesh (E, array).}
  Modality array: a 15-dimensional magnetic vector per finger (5 magnets $\times$ 3 axes).
  MLP adapter $15 \!\to\! 64 \!\to\! 128 \!\to\! 512$ per finger with the same token injection as FlexiTac.
  The camera backbone learning rate is $10^{-5}$.
\end{itemize}

Vision-only ablations set tactile modality to none (cameras and proprioception only), using the same task datasets with tactile channels ignored.

\subsection{Latent tactile policies (\textsc{TactX})}
\label{app:policy_latent}

A frozen cross-modal VAE encoder maps each sensor's raw tactile observations to a shared 16-dimensional latent mean vector, $\mu$, offline before ACT training. Daimon latents are computed from tactile RGB composites, eFlesh latents from 15-dimensional magnetic readings, and FlexiTac latents from $12 \times 16$ resistive tactile grids. These precomputed latents replace the raw tactile inputs in the policy datasets, and the encoder remains frozen during policy learning. For each dataset, the latent representations are normalized using the corresponding latent mean and standard deviation statistics before ACT training.

A lightweight MLP adapter is applied independently to each finger:
\[ 16 \rightarrow 64 \rightarrow 128 \rightarrow 512, \]
mapping each \textsc{TactX} latent into a single tactile token for the transformer. Consequently, ACT does not require a tactile CNN encoder. We further reduce $\lambda_{\mathrm{KL}}$ from 10 to 1 to mitigate mode collapse on the relatively small per-task datasets. During deployment, only the sensor-specific encoder branch changes across tactile hardware, while the ACT policy weights and the shared 16-dimensional latent space remain fixed.

\subsection{Binary contact policies}
\label{app:policy_contact}

Contact policies encode a single binary contact bit per finger.
Offline, each timestep is labeled
$\mathbb{1}\!\left[\mathrm{mean}\!\left(|x - x_{\mathrm{baseline}}|\right) > \tau_s\right]$
using a per-sensor baseline $x_{\mathrm{baseline}}$ and threshold $\tau_s$ fit on training data (Daimon uses depth and shear channels before scoring).
ACT uses modality array with $tactile\_dim = 1$ and MLP adapter
$1 \!\to\! 64 \!\to\! 128 \!\to\! 512$.
The hidden layers $[64, 128]$ match the \textsc{TactX} run so that only input dimension (1 vs.\ 16) differs between contact and latent policies.
We set $\lambda_{\mathrm{KL}} = 1$, as for latent tactile.
The same ACT checkpoint can be served on different tactile hardware by swapping only $(x_{\mathrm{baseline}}, \tau_s)$ at inference.

\begin{table}[t]
\centering
\caption{Tactile-specific ACT settings. Shared backbone hyperparameters are in Table~\ref{tab:app_act_shared}.}
\label{tab:app_act_tactile}
\begin{tabular}{lcccc}
\toprule
Variant & Modality & Per-finger input & Tactile encoder in ACT \\
\midrule
Raw Daimon (D)      & image & $240{\times}320{\times}3$ composite & ResNet-18 $\to$ spatial tokens  \\
Raw FlexiTac (F)    & array & 192-D resistive grid               & MLP $192{\to}64{\to}128{\to}512$  \\
Raw eFlesh (E)      & array & 15-D magnetic vector               & MLP $15{\to}64{\to}128{\to}512$   \\
Latent \textsc{TactX} & array & 16-D $\mu$ (frozen VAE, offline)  & MLP $16{\to}64{\to}128{\to}512$    \\
Binary contact      & array & 1-D thresholded bit (offline)      & MLP $1{\to}64{\to}128{\to}512$    \\
\bottomrule
\end{tabular}
\end{table}

\section{Full Policy Evaluation Results}
\label{sec:appendix_policy_results}

\subsection{In-domain Policy Evaluation}
\label{sec:appendix_indomain_policy}

We report the full in-domain policy results for each tactile sensor and task in Table~\ref{tab:in_domain}.
These experiments evaluate whether tactile-conditioned ACT policies benefit from tactile input when trained and evaluated with the same sensor.
Raw tactile observations provide a strong same-sensor baseline, while \textsc{TactX} tests whether the shared latent preserves enough task-relevant contact information for downstream control.

\begin{table}[tb]
\centering
\footnotesize
\setlength{\tabcolsep}{4pt}
\renewcommand{\arraystretch}{1.1}
\caption{\textbf{In-domain policy performance.} Same-sensor train and test. \textsc{TactX} matches or exceeds both baselines on pick-and-place and recovers most of the in-domain insertion gain provided by raw tactile. Success rates over 10 rollouts; bold marks the best per task per sensor.}
\label{tab:in_domain}
\begin{tabular}{l l ccccc}
\toprule
\textbf{Method} & \textbf{Sensor} & \textbf{P\&P} & \textbf{P\&P (OOD)} & \textbf{Insertion} & \textbf{Wiping} & \textbf{Reorient.} \\
\midrule
\multirow{3}{*}{Vision only}
& Daimon   & 9/10           & 8/10           & 2/10         & 4/10 & 9/10 \\
& eFlesh   & 9/10           & 6/10           & 2/10           & 0/10 & 8/10 \\
& FlexiTac & 7/10           & 6/10           & 8/10         & 9/10 & 7/10 \\
\midrule
\multirow{3}{*}{Vision + Ground Truth Tactile}
& Daimon   & 9/10           & 7/10           & 7/10         & 10/10 & 8/10 \\
& eFlesh   & \textbf{10/10} & \textbf{9/10}  & 7/10           & 6/10 & 10/10 \\
& FlexiTac & 9/10           & 8/10           & 8/10           & 9/10 & 10/10\\
\midrule
\multirow{3}{*}{Vision + \textsc{TactX} (Ours)}
& Daimon   & \textbf{10/10} & 5/10           & 5/10           & 5/10 & 10/10 \\
& eFlesh   & \textbf{10/10} & 8/10           & 5/10         & 8/10 & 9/10 \\
& FlexiTac & \textbf{10/10} & \textbf{9/10}  & 8/10           & 9/10 & 10/10 \\
\bottomrule
\end{tabular}
\end{table}

\end{document}